\documentclass{article}
\usepackage{arxiv}
\usepackage[utf8]{inputenc} 
\usepackage[T1]{fontenc}    
\usepackage{hyperref}       
\usepackage{url}            
\usepackage{booktabs}       
\usepackage{amsfonts}       
\usepackage{nicefrac}       
\usepackage{microtype}      
\usepackage{lipsum}
\usepackage{graphicx}
\graphicspath{ {./images/} }
\usepackage{algorithm}
\usepackage{algpseudocode}
\usepackage{amsmath}
\usepackage{multirow}
\usepackage{tabularx, makecell, booktabs}

\newcommand\nl{\newline}
\usepackage{float}
\restylefloat{table}
\usepackage{booktabs}
\newcommand\myVSpace[1][10pt]{\rule[\normalbaselineskip]{0pt}{#1}}

\title{Machine Learning Challenges and Opportunities in the African Agricultural Sector.\\ A General Perspective.
}

\author{
 Racine Ly \\
  AKADEMIYA2063\\
  Kigali, Rwanda \\
  \texttt{rly@akademiya2063.org}

}

\makeatletter
\renewcommand\@biblabel[1]{}
\renewenvironment{thebibliography}[1]
     {\section*{\refname}%
      \@mkboth{\MakeUppercase\refname}{\MakeUppercase\refname}%
      \list{}%
           {\leftmargin0pt
            \@openbib@code
            \usecounter{enumiv}}%
      \sloppy
      \clubpenalty4000
      \@clubpenalty \clubpenalty
      \widowpenalty4000%
      \sfcode`\.\@m}
     {\def\@noitemerr
       {\@latex@warning{Empty `thebibliography' environment}}%
      \endlist}
\makeatother

\begin{document}
\maketitle



\section{Introduction}
The improvement of computers’ capacities, advancements in algorithmic techniques, and the significant increase of available data have enabled the recent developments of Artificial Intelligence (AI) technology. One of its branches, called Machine Learning (ML), has shown strong capacities in mimicking characteristics attributed to human intelligence, such as vision, speech, and problem-solving. However, as previous technological revolutions suggest, their most significant impacts could be mostly expected on other sectors that were not traditional users of that technology. The agricultural sector is vital for African economies; improving yields, mitigating losses, and effective management of natural resources are crucial in a climate change era. Machine Learning is a technology with an added value in making predictions, hence the potential to reduce uncertainties and risk across sectors, in this case, the agricultural sector. The purpose of this paper is to contextualize and discuss barriers to ML-based solutions for African agriculture. In the second section, we provided an overview of ML technology from a historical and technical perspective and its main driving force. In the third section, we provided a brief review of the current use of ML in agriculture. Finally, in section 4, we discuss ML growing interest in Africa and the potential barriers to creating and using ML-based solutions in the agricultural sector.

\section{Machine Learning: An Overview}
\textbf{A historical perspective of Machine Learning}

Machine Learning is not a recent area of research, and its developments can be traced back to advancements in modeling brain cell interactions. In his book, The Organization of Behavior (Hebb, 1949) introduced a model of excitements and communications between neurons as entities that are interconnected and capable of processing and firing information to other neurons. In the same period, in 1950, the gaming industry proposed an algorithm for playing checkers. The innovation was the algorithm being equipped with a minimax reward function that scored each move and measured each side winning chance (Samuel, 1959). Those two historical advancements paved the way for Artificial Neural Networks (ANNs), with both ideas combined in creating the perceptron by Frank Rosenblatt (Rosenblatt, 1958). The perceptron is a mix of Donald Hebb’s brain cells interaction model and advancements in learning machines from Arthur Samuel’s work. Since then, the ML community has discovered that a multi-layer perceptron can yield better accuracy results. Such advancements also lead to a plethora of architecture to improve machine learning capabilities; The most famous is the feed-forward neural network. \\
\newline
The most impactful advancement that propelled most ML research areas’ developments is the backpropagation algorithm (Rumelhart et al., 1986). Backpropagation is a generalization of optimization algorithms for Artificial Neural Networks. As with other gradient-descent-based algorithms, its purpose is to update parameters of a chosen hypothesis that is expected to mimic the data patterns by minimizing a cost function, which assesses how close the hypothesis is to the actual data given certain mathematical constraints.  While researchers have long been familiar with the sequential and iterative aspects of such algorithms, the backpropagation algorithm added value is how it has been adapted to ANNs architecture to update neuronal connections’ weights by back-propagating residuals between predicted and actual values. Nowadays, several ML techniques are being developed for different applications. The main advancements are driven by the pursuit of faster learning processes (Castillo et al., 2006; Fu, 2017), more accurate predictions (Brohi et al., 2019; González-Sanchez et al., 2014), and the embedment of such algorithms into devices.\\

\textbf{Machine Learning Techniques}

Machines can be trained in different ways depending on the type of data and the application. The overarching goal is for the machine to learn the data patterns and to make future predictions. The three most used learning processes are the supervised, unsupervised, and reinforcement learning techniques.

\begin{itemize}
\item \textbf{Supervised learning} \\
Supervised learning is used when explanatory variables and their corresponding labels are available (Liakos et al., 2018). The machine is set to learn the interrelationships between them and generate predictions based on new inputs. Most of the time, the challenge is to find an equilibrium between bias and variance. A biased ML model refers to an underfitting problem where the algorithm did not learn enough from the data. The representation of what the data patterns look like is inadequate, and predictions are not accurate. On the other hand, a ML model with a variance problem refers to an overfitting problem. The algorithm is overperforming on mimicking the data, and the outcome is a lack of good generalization on new input data. 

\item \textbf{Unsupervised learning} \\
On the contrary of supervised learning, unsupervised learning is used when the explanatory variables’ labels are not available. The goal is to let the machine learns from the data by identifying their similarities. The outcomes are clusters representing groups of data points that the machine believes are closed in distance. Unsupervised learning is less used and monetized than supervised learning, but it is usually used as a first step to build label datasets. The main challenge of unsupervised learning is the optimal number of clusters to be set for the algorithm. Notwithstanding the development of techniques such as the “Elbow method” (Thorndike, 1953), human expertise is sometimes still needed to assess it.

\item \textbf{Reinforcement learning} \\
Reinforcement learning is operating differently compared to the techniques mentioned above. Supervised and unsupervised learning techniques use an already built dataset, while reinforcement learning builds the dataset along the learning process. The learning part follows a stage where the machine conducts a specific task on its own. When the outcomes are desired, the machine receives a reward, and a penalty is provided otherwise. Subsequently, the machine will reinforce the patterns that yield the expected outcomes. Such methods are primarily used in robotics (Kober et al., 2012) and the gaming industry (Silver et al., 2017).

\end{itemize}

\textbf{Machine Learning and traditional regression techniques}

Machine Learning is different from traditional regression techniques. Regression is a statistical method that tries to make predictions based on the average of past values. The keyword is average since its constitutive equations are meant to reduce prediction errors in average, which can also be translated into missing the actual value at each prediction. Machine Learning algorithms tend to be less performant in predicting values on average. At the same time, they can yield good accuracy with the actual targets: predictions can be biased in exchange for reducing variance. In addition, the way both regression and ML techniques are being built can explain how fast the latter evolves. To be adopted, regression techniques usually need to be theoretically proven, while ML practitioners rely on experiments to show how well their models perform on specific tasks. The tendency for experiments allows a faster evolution for ML compared to regression techniques. ML algorithms led to failures and disappointments from initial promises to revolutionize prediction technologies at their early stages. The systematic comparison between traditional regression-based techniques and ML results showed that it was premature at the time to assume that the algorithms would outperform statistical methods. However, with improvements in mathematical modeling, data availability, and greater computing capacities, ML algorithms now yield satisfactory accuracy in some areas, even though regression-based techniques can still provide better results in specific contexts (Ly et al., 2021).

\textbf{Machine Learning Driving Force}

As the Internet revolution reduces the cost of information gathering, ML reduces the cost of making predictions and enhances its utilization in many other areas. Weather, stock exchange, and commodity prices, among others, are domains that are traditional users of predictions. The capacity to reframe a problem from a non-traditional prediction sector into a prediction problem, a process we call AI insight, is one of the fuels of ML developments. Engineers have long been designing autonomous cars, but in a well-controlled environment and with explicit programming techniques. The shift towards ML-driven autonomous cars happened when the following question was raised: What would a human do? This type of question is one of the driving forces behind the transformation of many sectors working on implementing or already using ML in their businesses.

\section{Machine Learning use in the Agricultural Sector}
In the literature, the interrelationships between structural transformation, economic growth, and agricultural development have long been debated, and the latter has been neglected in theories of development. In his paper (Rosestein-Rodan, 1943), stated that: “The main role of industrialization is … to transform … peasants into full or part-time industrial workers”, and (Lewis, 1954), with his two sectors model considered the labor coming out of the Agricultural sector is an unlimited resource. The above treatment of the agricultural sector implies that agricultural growth is not affected by industrialization and has no significant impact on that process (Badiane, O., 1990). Since then, many other re-assessments have paved the way to reintroduce the agricultural sector into the scheme (Badiane, O., 1990). Industrialization is the ultimate engine for economic growth, and the development of the agricultural sector is a good long-term strategy for industrialization. In this section, we discuss how ML can be harnessed for the development of the African agricultural sector.\\
\newline
Agriculture has long been a laboratory for testing new technologies for many reasons: 

\begin{itemize}
\item The worldwide need to increase agricultural production for consumption. 
\item We live in the automation era and its potential for income growth and wealth creation for farmers, agricultural value chain, and non-agricultural actors. 
\item The lack of information on markets, weather, and potential crop growing conditions evolution.
\item The agricultural sector is an employer of a large portion of the population, therefore a suitable environment for technologies' scalability testing. 
\item The inevitable need for optimizing agricultural operations for resources’ management, climate resilience, pesticides and fertilizers rationalized use. 

\end{itemize}

Most of the aspects mentioned above have been successfully tackled with digital tools. Each of them was ushered through greater technological adoption: mechanization, biological engineering, digital-enabled tools, applied mathematics techniques, semiconductors and chemicals, and satellites and drones. Their common characteristic resides in being explicitly programmed by humans with rules to follow in specific situations. However, despite their effectiveness in solving specific problems within the agricultural sector, they are usually not interoperable as a system.\\
\newline
Machine Learning seems to spread faster into sectors than previous technological advancements since the ML insight could be applied everywhere. Reframing rule-based processes into experience-based algorithms has been enabled by the long “enough” data gathering period that came after each revolution mentioned in the previous paragraph. The impacts of ML on the agricultural sector are not limited to one section of the food system, but most of them: farm operations, seed selection, planting periods, market prices, distribution, among others. Previous digital technologies have been able to bring features that are mainly aimed at filling the gap in organizational skills: planning tools and dashboards, tracking devices. However, the ML “revolution” should not only be expected in those areas but also in being able to make predictions and prescriptions to reduce risks and uncertainties in decision-making processes, allowing farmers and other actors for early planning.\\
\newline
The United Nations stated that Africa would be home to approximately 1.68 billion people by 2030, which corresponds to an increase of +42\% compared to the African population in 2015 (United Nations, 2015). Therefore, food production for consumption should be increased. Simultaneously, efforts should also be made to ensure adequate nutritional content and good practices for environmental sustainability. However, African farmers are mostly smallholders: 95\% of African farm sizes are less than 5 hectares (Lowder et al., 2014). Land extension for more production is usually tricky since ownership could be informal and unclear. An alternative to increase agricultural yield could be optimizing farm operations and reducing what we call hidden losses. In a 5-ha agricultural land, several aspects are subject to uncertainties that can lead to poor performance: erratic rainfall, lack of knowledge about biophysical parameters, soil water content, inadequate planting period. Technology advancements are suitable to tackle those issues: Indeed, the Internet of Things, satellite images, and drones have shown their positive impacts in providing measurements of those parameters.\\
\newline
Sensing technologies have significant monitoring capability, which is of great importance in assessing trends. However, they tend not to be suitable in predicting corresponding values in the near future to mitigate the risks mentioned above. In addition, they only measure one or two parameters, while the underlying factors of poor performance can be multivariate. This is where prediction is most needed: Machine Learning is a prediction technology that can identify the most likely essential inputs within the training set features for better predictions. Several inputs of different nature can be combined in predicting a variable that helps to be closer to reality, contrary to mechanistic approaches, which rely on mathematical descriptions of biophysical processes. 

\textbf{Agricultural yield prediction, diseases detection, and feature recognition}

A large number of ML-based applications are being harnessed in the Agricultural sector. For yield prediction, (Kaneko et al., 2019) proposed a deep long-short term memory model to predict maize yield in six African countries: Ethiopia, Kenya, Malawi, Nigeria, Tanzania, and Zambia. Their input data are derived from MODIS multispectral products, and they applied the data dimensionality reduction technique used in (You et al., 2017). They have been able to predict maize production at county levels accurately and discussed that training all countries simultaneously is a promising artifact in improving prediction accuracy for those who suffer from data scarcity. In (Ly and Dia, 2020), the authors developed a food crop production model that uses MODIS biogeophysical parameters as inputs. The model provides production forecasts at a grid cell size of ten kilometers which corresponds to the community level for most African countries. In (Ferentinos, 2018), a convolutional neural network model was developed to detect plant diseases from RGB leaves pictures. The training dataset contains more than 87,000 images for 25 different plants in a set of 58 distinct classes of (plant, disease). The overall accuracy of 99.53\% has been performed, making the model an interesting tool for early warning systems for farmers. Similar techniques have been used for weed detection (Dyrmann, 2018), species recognition (Abdullahi et al., 2017; Zhang and Zhang, 2017; Bao et al., 2019), tree counting (Cheang et al., 2017), and crop quality assessment (Chokey and Jain, 2019). 

\textbf{Rainfall and Soil Moisture}

Water is a valuable asset and a mandatory need for agriculture. There are two main ways to access and use water in agriculture: rainfall and irrigation. Having access to current and future water availability information is a game-changer for African farmers to reduce their vulnerability in the context of climate change with erratic water cycles. In the literature, several ML models are being harnessed to address that knowledge gap. (Oswal, 2019) proposed a binary classification model built with logistic regression technique to assess if it will rain tomorrow or not in Australian cities, though without providing information about its intensity. (Hernandez et al., 2016) Columbia also followed the same objective by using a deep learning approach, and their results outperform other known models that use classic neural network architecture. Even most of the available ML-based models concern only small areas and still need to be scaled up, such rainfall information is crucial for smallholder farmers. One might plan to sow the land at a period that does not match the erratic raining patterns; Therefore, the entire production would be affected. When weather seasonality is extending or shortening, adaptation is critical to sustaining at least reasonable agricultural yields, if not improvements.\\
\newline
Soil moisture is also one of the main factors in agricultural production and hydrological cycles, and its prediction is essential for the rational use and management of water resources. In Africa, water scarcity is problematic for smallholder farmers (Giordano et al., 2019), and access to irrigation facilities is difficult (Mwamakamba et al., 2017). Information about soil moisture is paramount to assess whether the soil needs water or not. Most African agriculture is rainfed (Abrams, 2018); therefore, soil moisture content information is needed for water-saving and, most and foremost, in adapting farmers’ habits in terms of water use in their daily activities on the farm. Several soil moisture content prediction models are being built around the world using ML techniques. (Cai et al., 2019), proposed a deep learning model that can predict soil moisture content in Beijing; (Prakash et al., 2018) also compared several ML-based models to predict soil water content for the next 1, 2, and 7 days. In their paper, it can be found that the maximum accuracy measured through the Root Mean Squared Error (0.975) is reached when predicting for the next day. When irrigation is available, water management can also be performed through irrigation scheduling systems. (Adeyemi et al., 2018) developed a Long Short-Term Memory model to predict soil moisture content. The predicted values are fed into a rule-based irrigation system that can operate when a volumetric threshold is reached. The model is developed using the Cosmic-Ray Soil Moisture Observing System (COSMOS), a project led in the United Kingdom and operating in the UK, USA, Australia, and China (Zedra et al., 2012). (Zhang et al., 2018) also proposed an extended short-term memory model to predict water table depth in agricultural areas with monthly water diversion, evaporation, precipitation, temperature, and time as input data. The results showed a much higher R2 score compared to Artificial Neural network approaches. 

\textbf{Market price forecasting}

While mitigating hidden losses is more focused on on-farm activities and agricultural yield growth, developing the agricultural sector also suggests increasing farmers’ incomes through better integration into the agricultural value chain. The purpose of increasing farmers’ incomes is to reduce poverty and allow them to provide greater access to healthcare, education, and nutritious food for their families. It hopefully also means more financial resources to be invested on the farm in terms of improved seeds, mechanics, irrigation, and machines for agricultural product transformation. In fulfilling such ambition, selling agricultural products at optimal price is of great importance. (Ouyang et al., 2018) proposed using a long-short term memory model to predict commodity prices with high accuracy, and such information could close the knowledge gap for farmers about market prices.\\
\newline
Other applications are being built for automation into the agricultural sector, including AI-driven tractors and harvesters see (Liakos, 2018) for a general literature review of harnessed ML techniques in agriculture. Despite ML can improve agricultural yields through farm operations’ optimization, reduce hidden losses and uncertainties through predictions, shorten decision-making processes, its adoption into Africa’s agricultural sector is still slow. Beyond the condition that it has to be a cost-reducing technology for farmers to enhance that adoption, other factors could also be operating. The new wave of ML revolution is still nascent, and the growing interest in African countries follows a bottom-up streamline in terms of policy framework. Identifying barriers to adopting ML-based solutions in Africa’s agricultural sector could help target investments and policy design.

\section{Machine Learning Growing Interest in Africa and Adoption Barriers}
\textbf{Machine Learning interest in Africa}

A growing interest in Machine Learning is undergoing around the world, and a good proxy for that is the investments in AI start-ups. Most of the AI-stat-ups use ML techniques as the technology that drives their businesses. In 2017, USD 15B of total investments had been injected into AI start-ups worldwide, with a cumulative amount of USD 50B since 2011 (OECD, 2018 - Figure 1). The United States of America is the biggest investor in AI, followed by China and the European Union. Those investments are led mainly by technological firms that acquire promising start-ups. In the non-technological sector, AI adoption is at its early stages; a few firms have deployed AI solutions at scale.

\begin{figure}[htbp!]
  \caption{Total estimated investments in AI start-ups, 2011-2017 and first semester 2018. - OECD estimation, based on Crunchbase (July 2018).}
  \centering
    \includegraphics[width=1\textwidth]{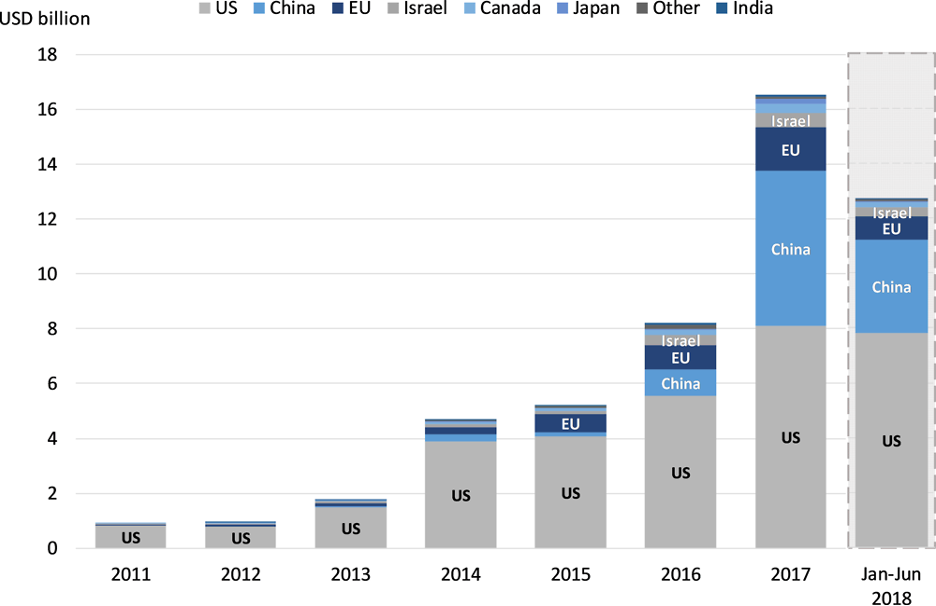}
\end{figure}

\begin{figure}[htbp!]
  \caption{(a) Number of companies using Artificial Intelligence in Africa - Q3 2019. Source: Briterbridges report on Adopting Artificial Intelligence in Africa - Quarter 3 2019 in collaboration with Alliance AI. 2. (b) AI readiness Index. Data source: Oxford Insights. Maps produced by the author.}
  \centering
    \includegraphics[width=1\textwidth]{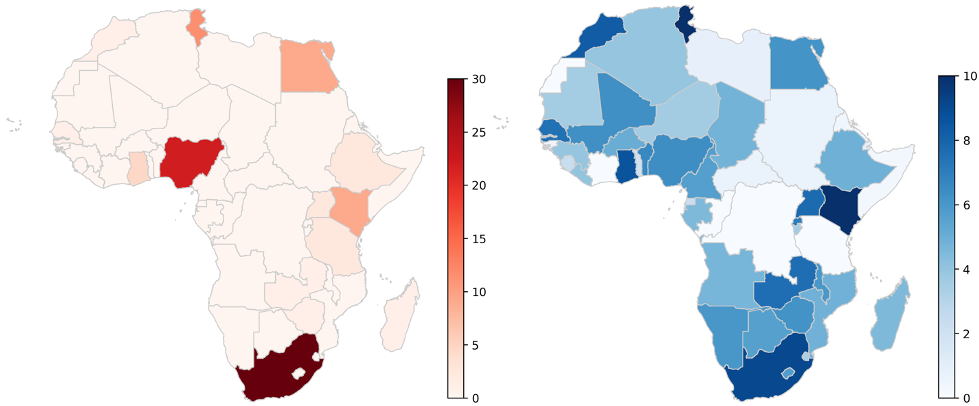}
\end{figure}

There is also a growing interest in Artificial Intelligence in Africa, even though the lack of records makes the monitoring process difficult. Research papers and AI adoption reports assess how AI can be used to solve the most pressing issues in Africa (Elisa, 2018), the number of start-ups claiming to use AI, and the AI-induced growth for African countries’ GDP. During the third quarter of 2019, 86 start-ups have been listed as AI-driven, with South Africa, Nigeria, and Tunisia at the top three positions. They are counting 26, 19, and 10 start-ups, while other countries have between 0 and 8. Figure 2. (a) shows a nascent ecosystem to drive the AI industry in Africa. It would also be correct to say that the number of AI-driven companies might not be the only proxy to look at for several reasons. Machine Learning, as the main driver of the AI industry, is now a democratized knowledge with many online materials to learn its basics and more advanced courses. Until a recent past, one might need to enroll in an academic program to access that knowledge; today, a simple registration to an online course can grant access to that expertise. The growing community of machine learners maintains the pace of knowledge-sharing and online collaboration. In addition, Africa is becoming a hub for technology companies known for Machine Learning enablers such as Google, IBM, and Microsoft. Therefore, their presence could enhance the entrepreneurship ecosystem and public-private partnerships, potentially impacting public awareness of such technologies. \\
\newline
Another reason why Africa’s AI industry is lagging compared to other regions of the world is its informal aspect. Many initiatives across the continent are not being documented because they partly do not lead to company creation. Africa is constituted of developing countries; hence imperfect capital markets, which are a crucial determinant of informality, could be a barrier for formal entrepreneurship and the international visibility which comes with it (Straub, 2005; De Mel et al., 2008; Grimm et al. 2010). Another critical determinant of informality is entry sunk costs to the formal sector, which are proportionally higher in developing countries than in advanced economies. Similarly, excessive or inappropriate government regulations have also been a significant constraint to entrepreneurship (Auriol E., 2014).\\
\newline
Potential AI-induced GDP growth of +5.6\% (\$1.2 trillion) could be expected in Africa, Oceania, and other Asian markets by 2030 (PWC report, 2017). Such should raise the question of the preparedness of African countries to take most of the AI revolution. By comparing figures 2.a and 2.b, two observations can be made: First, the top-ranked countries in terms of the number of AI-driven companies are part of the readiest to thrive into the AI ecosystem. Second, not all the African countries with the highest AI readiness fulfill that potential, therefore indicating barriers that need to be overcome for a full deployment of AI (through ML as its driving technology) into different sectors, including agriculture.

\textbf{Potential Machine Learning Production Barriers in Africa’s Agricultural Sector}

Assessing Machine Learning penetration into Africa’s agriculture can be difficult due to the lack of information about the sector. However, since more interest is being noticed across the continent, more data should be expected about ML adoption by farmers and agribusinesses. In this section, we discuss potential ML production barriers in Africa’s agricultural sector. We define Machine Learning production barriers as the existing gap for a country to fulfill its full Machine Learning potential in producing Machine Learning solutions for farmers and agribusinesses. Analyzing countries’ Machine Learning potential versus the number of ML solutions for agriculture should depict a country ML production barrier profile. Though, two research hypotheses have been made: i) We consider ML as the driving force of the artificial intelligence revolution; ii) We consider that any existing African technology company operating in the agricultural sector, and claiming to use AI, have at least one commercial product on the market for farmers and agribusinesses.\\
\newline
We used an Oxford Insights and the International Development Research Center dataset, which consists of eleven parameters grouped into four clusters: Governance, Infrastructure and Data, Skills and Education, and Government and public services (Oxford Insights, 2019). One hundred ninety-four countries have been ranked for each parameter, and a global AI readiness index is attributed to each of them. Table 1 gives a general overview of indicators, clusters, and data sources used to compute the index. Each indicator value is normalized with its maximum value when all countries are taken into account. Countries’ normalized scores for each indicator are summed to give the AI readiness index between 0 and 11, the latter being the maximum state of readiness according to the methodology.

\begin{table}[H]
\caption {Clusters, Indicators, and Data sources used for the AI readiness index computations.} \label{tab:table1} 
    \begin{tabularx}{\linewidth}{l*{2}{X}c}
    \toprule
    \thead{Cluster} & \thead{Indicators} & \thead{Data Source} \\
    \midrule
    Governance & Privacy law & UN Data protection and privacy legislation \\ \myVSpace[10pt]
    & AI strategy & Medium Article \\ 
    \midrule
    Infrastructure and Data & Data availability	& OKFN Open Data Index \\ \myVSpace[10pt]
    & Government procurement and advanced technology products & Government procurement of advanced technology products \\ 
    & Data capability & UN eGovernment Index \\
    \midrule
    Skills and Education & Technology skills & WEF global competitiveness report 2018 \\ \myVSpace[10pt]
    & AI Startups & Crunchbase \\ \myVSpace[10pt] 
    & Private sector innovation capability	& WEF global competitiveness report 2018 \\
    \midrule
    Government and Public Services & Digital public services &	UN online service index from UN eGovernment Survey \\ \myVSpace[10pt]
    & Effectiveness of government & World Bank 2017, Government effectivenes\\ \myVSpace[10pt]
    & Importance of ICTs to government vision of the future & WEF Networked Readiness Index 2016
    \hrule
    \end{tabularx}
\end{table}

In addition, the number of African technological companies operating in the agricultural sector and claiming to use artificial intelligence has been extracted from (Briterbridges, 2019). The picture is clear: Africa, given its potential through the governments’ AI readiness indicator, does not have enough AI-driven companies to impact other sectors such as agriculture. There are no African countries in the top 50 positions, and only 12 African countries (out of 54 in the list) are in the top 100. The top five African governments - Kenya, Tunisia, Mauritius, South Africa, and Ghana, reflect their well-documented technology advancements. Among the last ten ranked African countries, seven are least developed countries. \\
\newline
Even though data capabilities seem to be strong (figure 3.E) for most African countries, technological skills status is high for many African countries (figure 3.F), and governments believe that ICTs are essential for the future (figure 3.K), such observations cannot be reflected in the number of AI start-ups creation (figure 3.G). Therefore, in terms of Machine Learning production barriers for the agricultural sector, the main issues should be investigated on both sides: governments and entrepreneurs. 

\begin{figure}[htbp!]
  \caption{Governments’ AI readiness index sub-indicators mapping. (C): Data availability, (D): Government procurement and advanced technology products, (E): Data capability, (F): Technology skills, (G): AI Startups, (H): Private sector innovation capability, (I): Digital public services, (J): Effectiveness of government, (K): Importance of ICTs to government vision of the future}
  \centering
    \includegraphics[width=0.7\textwidth]{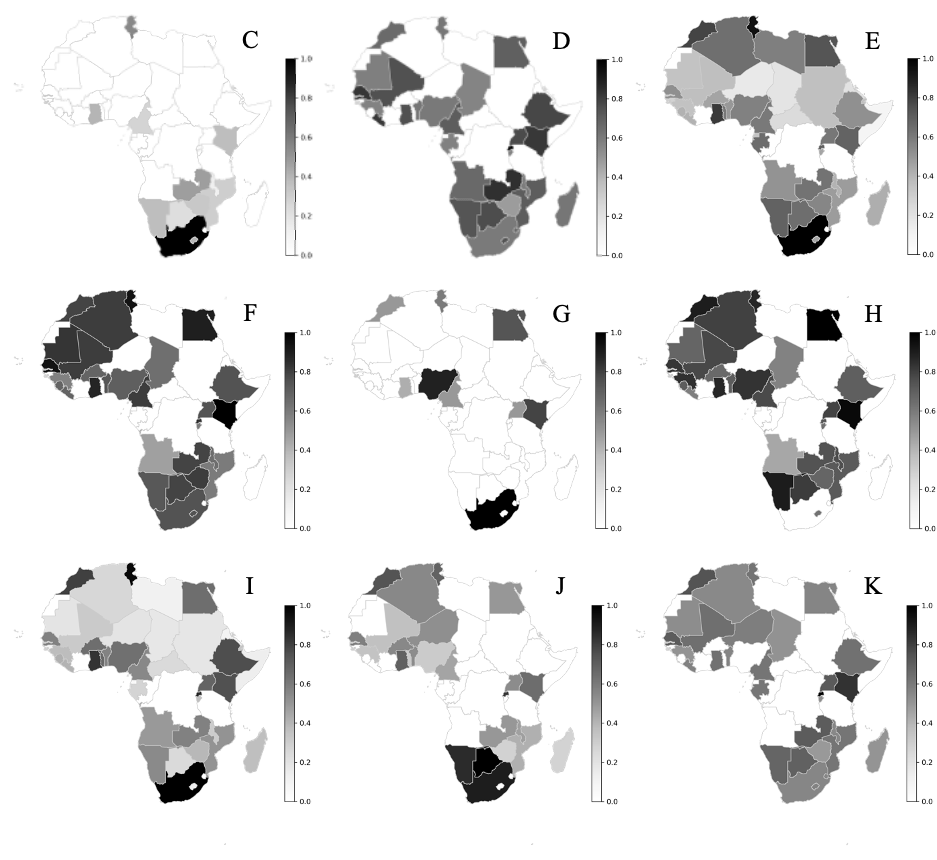}
\end{figure}

From the government side, only a few of them have an AI strategy that could set a framework to enhance its development. In the 54 countries encountered in the dataset, only Kenya and Tunisia have one. They also count eight and ten AI firms in general and two and one AI firms operating in the agricultural sector. Given such facts, it seems that having a national AI strategy is likely to set a favorable environment for entrepreneurship in AI first, and to some extent, to the agricultural sector. From the entrepreneurship perspective, only nine countries have AI firms that are operating in the agricultural sector: Egypt (1), Ethiopia (1), Kenya (2), Nigeria (1), Rwanda (1), Kenya (1), Tunisia (1), Uganda (1) and Tanzania (1). Several reasons can explain such low activity.

\textbf{Data availability}

Machine Learning is a data-consuming technology, and the lack of data is undoubtedly a strong barrier for Machine Learning solutions’ creation and deployment into the agricultural sector. Figure 3.C shows a score of zero in terms of data available for most African countries. It also means African countries with higher data availability scores make them more available to the public. In contrast, no distinction can be made for other countries between data availability and data scarcity.

\textbf{Data scarcity}

Governments need data for their policies’ design, decision-making, and monitoring and evaluation processes. Thus a particular type of data is being collected and used. However, a low score in data availability can primarily mean it is not shared with the public or data scarcity is operating. The latter for machine learning solutions for the agricultural sector is more about the quality and the quantity of the datasets. For ML to be used, large enough, long time series and disaggregated quality datasets are needed. A long-standing tradition of data gathering is paramount. In Africa, as shown in figure 3.I and 3.J, digital public services and government effectiveness are mitigated even though a digitalization process for surveys and similar public activities are being noticed relatively recently (World Bank Group, 2017). Data gathering might have been conducted in the past, but with other means than digital devices and cloud as a storage and computing facility (like paper forms), which suggest a higher risk of data losses. In addition, when efforts are made to gather better quality data and their availability, one possible barrier could be found in public awareness of the existence of such data.

\textbf{Complements of Machine Learning accessibility}

When the price of goods drops, we use more of it, and simultaneously the cost of a few related goods might increase; this is called complements. Machine Learning diminishes the cost of making predictions; therefore, the primary ingredient for doing Machine Learning, which is data, might increase. More specifically, the cost of devices that facilitate data gathering can be more expensive. In the agricultural sector, ground truth data are needed to make valuable predictions. One barrier for entrepreneurs to tackle agricultural sector issues could also come from a higher price of sensors to gather data and harness ML techniques for farmers and agribusinesses. The complements issues are more about sensors to use on the ground to measure soil water content, humidity, temperature, wind speed, solar radiation, image acquisition for image recognition, among others. Today, remote sensing through satellite images and drones is used to close the data gap in Africa, but specific skills are required.

\begin{figure}[htbp!]
  \caption{Governments’ AI Readiness sub-indicators correlation matrix. Data Source: Oxford Insight, Government AI readiness 2019. Figure source: Author. (A): Privacy law, (B): AI Strategy, (C): Data availability, (D): Government procurement and advanced technology products, (E): Data capability, (F): Technology skills, (G): AI Startups, (H): Private sector innovation capability, (I): Digital public services, (J): Effectiveness of government, (K): Importance of ICTs to government vision of the future.}
  \centering
    \includegraphics[width=0.7\textwidth]{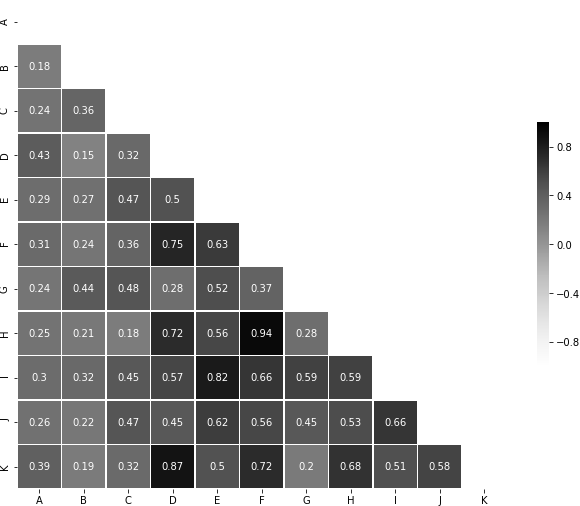}
\end{figure}

The development of digital public services (column I) and data capabilities (column E) are most likely to enhance the number of AI start-ups creation, which, by extension, will reach the agricultural sector because of its long tradition of being a laboratory for testing new technologies (Figure 4). However, machine learning is still nascent in Africa. A strategy to raise the level of expertise in the sector will create a critical mass of engineers and experts to lead the private sector. Figure 4 also shows a high correlation coefficient of +0.94 between the technological skills (F) and private sector innovation capabilities (H) sub-indicators, which can be considered a causality effect.

\begin{table}[htbp!]
\caption {Proposed sectors and corresponding actions for Machine Learning development in Africa’s Agricultural sector.} \label{tab:table2} 
\begin{tabularx}{\linewidth}{l*{2}{X}c}
\toprule
AI strategy & Entrepreneurship & Facilitate company creation to incentivize Agtech services and products providers that use ML with tax exoneration.\\ \myVSpace[10pt]

                 & Education and Research & Create attractive academic programs in data-related sciences, ML with a major in agricultural sciences.\nl 
                 \myVSpace[10pt] Fund research programs that use ML techniques for topics in the agricultural sector. \nl 
                 \myVSpace[10pt] National public funding for doctoral programs to train future experts in ML.\\ \myVSpace[10pt]
                 
                 & Data gathering facilities and data management systems & Invest in public cloud systems with high computational and storage capacities.\nl \myVSpace[10pt] Invest in digital data gathering tools such as tablets, sensors (solar energy), and drones.  \nl 
                 \myVSpace[10pt] Invest in a data management system to analyze and visualize data from devices.\\ \myVSpace[10pt]
                 
                 & Data availability and ethics  & Create national agencies for AI that support public services and researchers and making datasets publicly available.\nl 
                 \myVSpace[10pt] Observatory for Machine Learning societal bias and unethical use.\\ \myVSpace[10pt]
                 
                 & Privacy law  & Ensure that countries have privacy law that protects individuals and communities.\nl \myVSpace[10pt] Create committee that will be responsible for updating the privacy law regularly.\\
\midrule
Foster AI use in public services & & Modernize public services with the use of ML solutions. Such adoption in public services can foster its diffusion in other sectors.\\
\midrule
AI Forum & & An AI forum could be the place of high-level discussions and knowledge sharing between experts, researchers, policymakers, deciders, and private sector actors at a continental level. Such an event should also be the place to discuss coordination at a regional and continental level about ML development.\\
\bottomrule
\end{tabularx}
\end{table}

\section{Conclusion and Recommendations}
This paper introduced Machine Learning from historical and technical perspectives presented its differences with traditional regression techniques and highlighted its driving forces. We also discussed a rising interest in ML across Africa and its potential application in the agricultural sector. The previous section was about potential ML production barriers for agricultural development. In this section, we finally discuss a few recommendations for ML-based solutions’ creation and adoption.\\
\newline
While the private sector mainly drives ML solutions creation (therefore, its adoption could be following a bottom-up scheme), short- and long-term policies are mandatory to set a framework for its development and adoption in the agricultural sector and beyond. Two main policy typologies could be emphasized: policies that set the right environment to facilitate the goals mentioned above; and policies that help mitigate the pitfalls of machine learning and define rules and laws to protect individuals. The first type of policy framework mainly focuses on long-term investments in education and research. It facilitates ML-driven companies’ creation with tax incentives (among other artifacts) for those who focus on the agricultural sector. In parallel, investments are needed to increase digital data gathering and analytics capacities (using tablets and sensors, cloud systems for data storage and computational capacities, web-based platforms for data analysis and visualization) and data management infrastructures. For the second type of policy framework, the goal is to mitigate protection and privacy issues that come with data availability, computational capacities, and Machine Learning skills. \\
\newline
Widespread ML adoption in societies also comes with challenges that can be nefarious to individuals and communities across ethnic, cultural, gender, or socioeconomic categories. Since Machine Learning solutions are built from historical data, any bias embedded within the datasets will potentially be reproduced by the algorithms. In Africa, where land heritage is predominantly patriarchal for most ethnic groups, if banks and investors rely on machine learning algorithms (something that is already happening) built on top of several features, including gender to provide loans to farmers, female farmers could be marginalized by the system, and further accentuate gender gaps. In this context, male farmers are not given loans because of their gender per se, but men mainly populated the ML model's historical dataset. Even if the dataset was built with a reasonable ratio of both genders, enough scenarios with positive examples where female farmers have been attributed loans are still needed by the algorithm to learn from those examples. Instead, if there are negative examples, the results will be similar – to some extent – when men dominate the dataset sample. Such a scenario suggests that: Machine Learning should not be used everywhere in the absence of good data; quality data is not only about disaggregation and number of data points, but also, and in this context, the absence of bias. There is a need to develop mechanisms to monitor Machine Learning ethics. \\
\newline
In summary, investments and incentives to stimulate the private sector for machine learning solutions’ creation for the agricultural sector, increased focus on education and research, modernization of data-gathering facilities, improvements on data management skills, ethical considerations in machine learning use, and privacy law frameworks, are critical components of national AI strategies that most African countries do not yet have. Given Africa’s potential to get the most out of the Machine Learning revolution, deliberate steps must be taken to fulfill this potential. Table 2 presents a list of sectors and corresponding actions we propose to create a minimum viable environment for ML development in the Agricultural sector. Finally, we also emphasize that an annual and Africa-wide AI Forum should be created as a place of expression and knowledge sharing between experts, researchers, policymakers, and deciders to better coordinate Machine Learning development at the continental level.

\bibliographystyle{unsrt}  


\end{document}